\documentclass[10pt,journal,compsoc]{IEEEtran}
\usepackage{graphicx}
\usepackage{amsmath,amssymb} 
\usepackage{color}
\usepackage{soul}
\usepackage{amsfonts}
\usepackage[utf8]{inputenc}
\usepackage{algpseudocode}
\usepackage{blindtext}
\usepackage{graphicx}
\usepackage{tikz}
\usepackage{makecell}
\usepackage{lineno}
\usepackage{multirow}
\usepackage{fancyhdr}
\usepackage{amsthm}
\usepackage{textcomp}
\usepackage{amsmath}
\usepackage{mdwmath}
\usepackage{bm}
\usepackage{adjustbox}
\usepackage{lineno,hyperref}
\usepackage{physics}
\usepackage{amsmath}
\modulolinenumbers[5]
\usepackage{commath}
\usepackage{algorithm}
\usepackage{bm}
\usepackage{float}
\usepackage{soul}
\usepackage[sorting=none]{biblatex}
\begin{document}
\title{Ensemble approach for detection of depression using EEG features}

\author{Egils~Avots, Klāvs~Jermakovs, Maie~Bachmann, 
        Laura Paeske, Cagri Ozcinar,
        and Gholamreza~Anbarjafari 

\IEEEcompsocitemizethanks{\IEEEcompsocthanksitem E. Avots, K. Jermakovs, C. Ozcinar, and G. Anbarjafari are with iCV Lab, University of Tartu, Estonia.\protect\\
\IEEEcompsocthanksitem M. Bachmann, and L. Paeske are with Biosignal Processing Laboratory, Tallinn University of Technology, Estonia.\protect\\
\IEEEcompsocthanksitem G. Anbarjafari is also with Yildiz Technical University, Istanbul, Turkey.\protect\\
He is also with PwC Advisery, helsinki, Finland.\protect\\
E-mail: shb@icv.tuit.ut.ee}
}

\markboth{}
{Avots \MakeLowercase{\textit{et al.}}: Detection of long lasting effects of depression using EEG and machine learning }

\IEEEtitleabstractindextext{%
\begin{abstract}
Depression is a public health issue which severely affects one's well being and cause negative social and economic effect for society. To rise awareness of these problems, this publication aims to determine if long lasting effects of depression can be determined from electoencephalographic (EEG) signals. The article contains accuracy comparison for SVM, LDA, NB, kNN and D3 binary classifiers which were trained using linear (relative band powers, APV, SASI) and non-linear (HFD, LZC, DFA) EEG features. The age and gender matched dataset consisted of 10 healthy subjects and 10 subjects with depression diagnosis at some point in their lifetime. Several of the proposed feature selection and classifier combinations reached accuracy of 90\% where all models where evaluated using 10-fold cross validation and averaged over 100 repetitions with random sample permutations. 
\end{abstract}

\begin{IEEEkeywords}
Depression, electroencephalogram (EEG), feature extraction and selection, machine learning, ensemble learning. 
\end{IEEEkeywords}
}

\maketitle

\IEEEraisesectionheading{\section{Introduction}\label{sec:introduction}}
\IEEEPARstart{D}{epression} is a major public health problem, creating a significant burden throughout the world. The World Health Organization (WHO) has predicted that the most common cause of incapacity to work and disability will be cardiovascular diseases and the second most common cause will be depression \cite{murray1996global}. According to disability adjusted life years or illness, depression ranks first in European countries \cite{wittchen2011size}. In Europe, the largest aggregate study of the prevalence of mental disorders in the population shows that clinically significant depression has been experienced by an average of 6.9\% of the European population in the last 12 months \cite{wittchen2011size}.

Currently, the most common way to diagnose depression is an interview conducted by a medical professional. In many cases the interview is accompanied by a clinical questionnaire either assessed by a medical doctor as Hamilton Depression Rating Scale (HAM-D) or self-reported as Emotional State Questionnaire (EST-Q). There is already DSM-5 and this is not a test, it is used to identify and classify diseases - would leave this out and would write only about questionnaires \cite{lehman2000diagnostic} or Mini-Mental State Examination (MMSE) \cite{tombaugh1996mini} to establish diagnosis criteria. Other questionnaires such as Beck depression inventory (BDI) \cite{beck1996manual} and Hamilton Depression Rating Scale (HDRS) \cite{mowbray1972hamilton} are used for screening purposes.

Besides subjective clinical questionnaires, brain activity of the patients can be monitored objectively applying various imaging modalities like Computed Tomography (CT), functional Magnetic Resonance Imaging (fMRI), and electroencephalogram (EEG). Out of those examples, EEG stands out as the most simple and cost effective device, hence the field of detecting mental states and disorders using EEG is an actively researched field showing promising results \cite{bachmann2018methods,cai2018pervasive,mahato2019electroencephalogram,khosla2020comparative}.

This paper contains classification results obtained by using various linear and non-linear features and general insight in the feature calculation. The main contribution of the paper is in the use of feature selection and classifier configuration to improve classification accuracy.

\section{RELATED WORK}
According to de Aguiar Neto et al \cite{de2019depression} absolute and relative band powers and various other linear and also non-linear features described in this section have been recognised as promising biomarkers for characterizing a depressed brain.

Absolute band power and relative band power of EEG signal have been analysed with Separate three-way multivariate analysis of variance (MANOVA) and showed that relative beta power was greater in depressed patients than in controls at all electrode locations and increased absolute beta power for some of the electrode locations \cite{knott2001eeg}.

The use of Alpha Power Variability (APV) and Relative Gamma Power (RGP) was proposed by Bachmann et al \cite{bachmann2018methods}. While APV indicates the power and frequency variations in alpha band, RGP characterizes the high frequency components. The differences between the depressive and control groups appeared statistically significant in a number of EEG channels leading to a Linear Regression classification accuracy of 81\%.

Spectral Asymmetry Index (SASI) indicates the relative asymmetry between higher and lower frequency bands. According to Hinrikus et al \cite{hinrikus2009electroencephalographic} SASI values differed significantly in all channels between healthy and depressed patients. Single EEG channel analysis has already shown positive results in the detection of depression \cite{bachmann2017single, bachmann2018methods}.

The nonlinear Higuchi's Fractal Dimension (HFD) calculates the fractal dimension of a signal in the time domain \cite{higuchi1988approach}. Bachmann et al \cite{bachmann2013spectral} applied HFD method for EEG signals and evaluated using Student’s T-test for two-tailed distribution with two-sample unequal variance to find if a statistical difference exists between depressed and healthy subjects. The alterations were statistically significant in all EEG channels and indicated 94\% of subjects as depressive in the depressive group and HFD indicated 76\% of subjects as non-depressive in the control group.

The nonlinear Lempel Ziv Complexity (LZC), introduced by Lempel and Ziv \cite{lempel1976complexity} measures the complexity of a signal and has been successfully used on EEG signals for the detection of different mental states \cite{zhang2001eeg, kalev2015lempel}. 

Detrended Fluctuation Analysis (DFA) \cite{peng1994mosaic} which indicates long-time correlations of the signal was applied to evaluate EEG signals and revealed a statistically significant difference between healthy and depressive subjects \cite{bachmann2014detrended}. In addition, LDA reached a classification accuracy of 70.6\% and by combining DFA and SASI, LDA classification accuracy increased to 91.2\% \cite{bachmann2017single}.

A comprehensive study by Bachmann et al \cite{bachmann2018methods} shows the diagnostic potential for linear (SASI, APV, RGP) and non-linear (HFD, DFA, LZC) features to classify depression. Single channel classification with logistical regression achieved an accuracy of 81\% using APV or RGP measures. The combination of two linear measures, SASI and RGP, reached an accuracy of 88\% and by combining linear and non-linear measures a classification accuracy of 92\% was achieved \cite{bachmann2018methods}. 

\section{Experimental Setup}

\subsection{EEG recording procedure}
\begin{figure}[htp]
    \centering
    \includegraphics[width=5cm]{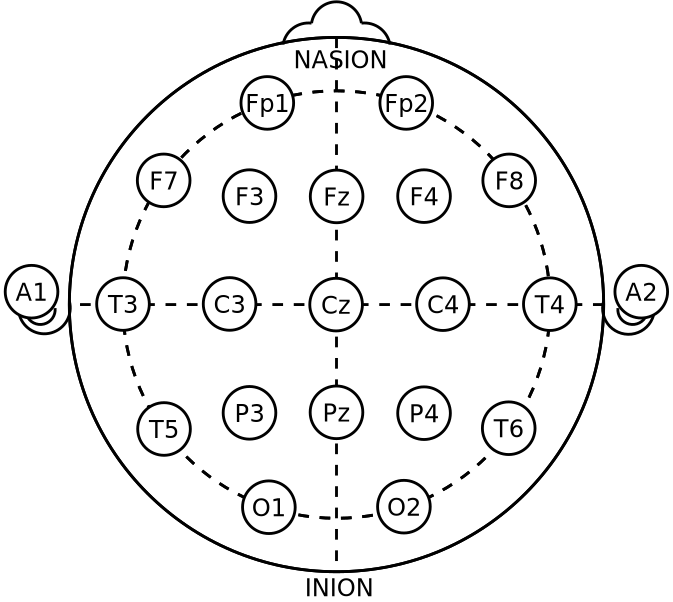}
    \caption{International 10-20 system for EEG recording.}
    \label{fig:sys}
\end{figure}

The Cadwell Easy II EEG (Kennewick, WA, USA) measurement equipment was used for EEG recordings using 18 channels (reference Cz), which were placed on the subject’s head according to the international 10–20 electrode position classification system shown in Fig. \ref{fig:sys}. During the recordings, subjects were lying in a relaxed position with their eyes closed. EEG signals within the frequency band of 3$-$48 Hz were used for further processing. The sample rate was kept at 400 Hz for linear methods, while the downsampled signals with a sample rate of 200 Hz were used for non-linear methods due to high computational load. The recorded EEG was segmented into 10-second segments and an experienced EEG specialist marked by visual inspection the first 30 artefact-free segments (5 minutes in total) for the subsequent calculation of features. 

The gathering of questionnaires and EEG recordings were carried out by the Tallinn University of Technology in accordance with the Declaration of Helsinki and the process was formally approved by the Tallinn Medical Research Ethics Committee. All participants signed the written informed consent. 

\subsection{Dataset}
The recorded dataset consisted of the EEG signals from 55 subjects regularly visiting the occupational health doctor of whom 20 were selected for further analyses, 14 females and 6 males within the age range from 24 to 53 years (mean age of 39 years old). Half of the subjects selected have been diagnosed with depression at some point in their life, while the age ($\pm$1.6 years) and gender-matched control group have never had depression diagnosis. In addition, the control group was chosen considering their low HAM-D and EEK-2 score to ensure they do not exhibit any signs of depression or other mental disorders.

\subsection{Hamilton Depression Rating Scale}
The HAM-D is the most widely used clinician-administered depression assessment scale. Although the rating scale has been criticized for use in clinical practice, in this study it was used as additional information for selecting healthy subjects. In situations where several healthy subjects were a match candidate for a depressive subject, the one with the lowest HAM-D score was chosen. The mean HAM-D score among healthy subjects was 3.1, where scores of 0-7 indicate no depression and a mean score of 9.3 for the depressive subjects which corresponds to mild depression. 

\subsection{Emotional State Questionnaire}
Emotional State Questionnaire (EST-Q) \cite{aluoja1999development} was originally compiled for use by lecturers of the psychiatric clinic of the University of Tartu in Estonia. The self-assessed questionnaire consists of 28 statements assessing major depressive and anxiety disorders and their associated symptoms during the last month. The questionnaire consists of 3 basic scales and 3 additional scales. Major scales include the depression, the general anxiety, and the panic agoraphobia subscale. Additional subscales include social anxiety, asthenia, and insomnia. Scale total score can be used as an overall indicator of the severity of emotional symptoms. The EST-Q was used in the current study for selecting healthy subjects to the control group. The subscale values of all the selected subjects were below the threshold, except for 3 subjects whose asthenia subscale was value.

\section{Features}
One of the most widely used method to analyze EEG data is to decompose the signal into functionally distinct frequency bands, such as delta (1-4 Hz), theta (4-8 Hz), alpha (8-12 Hz), beta (12-30 Hz) and gamma (30-45 Hz). Afterwards, additional information for each frequency band can be obtained, for example, average and relative band power and various linear and non-linear features.

\subsection{Relative band power}
One of the most widely used method to analyze EEG signals is to decompose the signal into functionally distinct frequency bands, such as delta (1-4 Hz), theta (4-8 Hz), alpha (8-12 Hz), beta (12-30 Hz) and gamma (30-45 Hz).
In current study this was achieved by first calculating the power spectral density of the EEG signal by Welch's method as by Bachmann et al., 2018. EEG powers in theta, alpha, beeta, and gamma frequency bands were computed integrating the power spectral density at the frequencies within the boundary frequencies of the EEG spectral bands.
In spectral analysis, it is common to take the magnitude-squared of the Fast Fourier Transform (FFT) to obtain an estimate of the power spectral density of the EEG signal. Afterwards, absolute band power can be obtained by calculating the area under curve using trapezoid rule for frequency bands. Relative band powers (T\textsubscript{RBP}, A\textsubscript{RBP}, B\textsubscript{RBP}, G\textsubscript{RBP}) are expressed as the power in the specific EEG frequency band as a percentage of the total power of the signal.

\subsection{Linear Features }

\subsubsection{Alpha power variability} 
The alpha band signal (8-12 Hz) was gained by pass-band filter. Next the APV was calculated to the artifact free 10-second segments in three steps. First, the alpha band signal power in time-window \(T\) for \(N = 4000 \) samples was calculated as
\begin{equation}
 W_i = \frac{1}{N}\sum_{r=1}^{N} [V(r)]^2
\end{equation}
where  \(V(r)\) is the amplitude of the alpha band signal in a sample \(r\) and \(N\) is the number of samples in the time-window \(T\). Afterwards, APV was calculated as
\begin{equation}
 APV = \frac{\sigma}{W_0}
\end{equation}
where \(W_0\) is the value of alpha band power averaged over 5 min and  \(\sigma \) is the standard deviation of those segments.

\subsubsection{Relative gamma power }
In recent studies brain activity in gamma range has been identified as a biomarker for major depression and gamma power in prefrontal cortex for stress assessment. RGP was calculated as fallows
\begin{equation}
 RGP_{mn} = R_{\gamma mn}/R_{\Sigma mn}
\end{equation}
where
\begin{equation}
 R_{\gamma mn} = \sum_{f_i=30}^{f_i=46} S_{mn}
\end{equation}
and 
\begin{equation}
 R_{\Sigma mn} = \sum_{f_i=3}^{f_i=46} S_{mn}
\end{equation}
where  \(S_{mn} \) is the power of the recorded EEG signal in a unit band at the frequency \(f_i \) in a channel \(m\) for a subject \(n\).

\subsubsection{Spectral asymmetry index}
SASI evaluates the power in higher and lower frequencies and was calculated as relative difference between the higher and the lower EEG frequency band power. The balance of the powers characterizes the EEG spectral asymmetry \cite{hinrikus2009electroencephalographic}. 
Powers in the frequency bands were calculated as
\begin{equation}
 P_{\delta mn} = \sum_{f_i=F_c-6}^{f_i=F_c-2} S_{mn}
\end{equation}
and
\begin{equation}
 P_{\beta mn} = \sum_{f_i=F_c+2}^{f_i=F_c+26} S_{mn}
\end{equation}
where \(F_c\) is the central frequency of EEG spectrum maximum in the alpha band and are calculated for each person individually.
SASI in channel \textit{m} for a subject \textit{n} is calculated as
\begin{equation}
 SASI_{mn} = \frac{ P_{\beta mn}-P_{\delta mn}}{P_{\beta mn}+P_{\delta mn}}
\end{equation}

\subsection{Non-Linear Features}
Non-linear methods are used to capture the chaotic behavior in EEG signal which happens due to underlying physiological activity occurring in the brain \cite{mahato2019detection}. To describe brain activity of the subjects we used Higuchi fractal dimension (HFD), Lempel Ziv complexity (LZC), and detrended fluctuation analysis (DFA).

\subsubsection{Higuchi Fractal Dimension}
Fractal dimension provides a measure of the complexity of time series such as the EEG and describes the fractal dimension of time series signals. The values of HFD for each electrode were calculated according to Higuchi \cite{higuchi1988approach} with parameter \(k_{max} = 8\).

\subsubsection{Lempel Ziv Complexity}

The complexity of signal can be quantified by LZC \cite{ibanez2015multiscale} describing spatio-temporal activity patterns in high-dimensionality nonlinear systems, it can reveal the regularity and randomness in EEG signals.
For LZC calculation each signal segment is converted into binary sequence \(s(n)\) as follows
\begin{equation}
s(n)=
\begin{cases}
  1, & \text{if}\ x(n) > m \\
  0, & \text{if}\ x(n) \leq m \\
\end{cases}
\end{equation}
where \(x(n)\) is the signal segment, \(n\) is the segment’s sample index from 1 to \(N\) (segment length) and \(m\) is the threshold value. The binary sequence \(s(n)\) is scanned from left to right counting the number of different patterns. The complexity value \(c(n)\) is increased every time a new pattern is encountered. LZC values are calculated as follows:
\begin{equation}
 C(N) = \frac{c(N)}{b(N)}
\end{equation}
where \(b(N)\) is the upper bound of \(c(n)\)
\begin{equation}
\lim_{n\to\infty} c(n)=b(N) = \frac{N}{log_aN}
\end{equation}
which is used to normalize LZC values to avoid variations in segment length.

\subsubsection{Detrended fluctuation analysis}
DFA is applied to evaluate the presence and persistence of long range correlations in time in EEG signals. It has been discovered that resting EEG of healthy subjects exhibit persistent long-range correlation in time \cite{bachmann2014detrended}.
DFA was calculated in the time domain according to the steps described by Peng et al \cite{peng1994mosaic,peng1995quantification}.

\section{Methodology}

The calculated features are represented as 1D vectors constructed from a set of feature calculations T\textsubscript{RBP}, A\textsubscript{RBP}, B\textsubscript{RBP}, G\textsubscript{RBP}, APV, RGP, SASI, HFD, LZC and DFA for electrodes FP1, FP2, F7, F3, FZ, F4, F8, T3, C3, C4, T4, T5, P3, PZ, P4, T6, O1, and O2. When analysed individually feature group consists of 18 features which represent the electrode locations. When using feature evaluation methods, such as F-test or ReliefF, the number of features can be reduced and such features can be concatenated together with other feature groups to build more diverse feature sets.

\begin{figure}[htp]
    \centering
    \includegraphics[width=9cm]{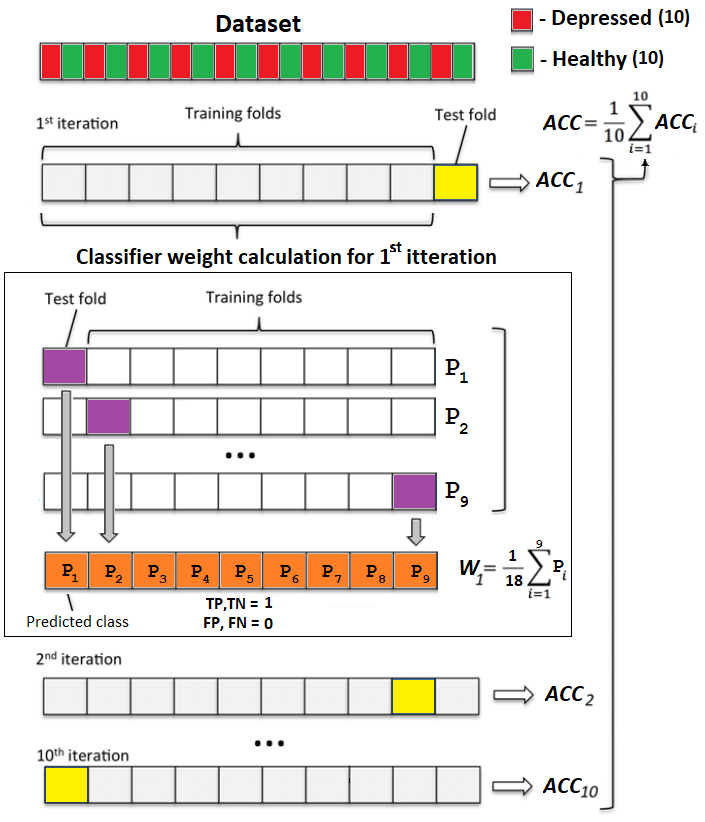}
    \caption{Cross validation process for ensemble classifier weight calculation}
    \label{fig:cv_chart}
\end{figure}

The individual feature groups, individual feature groups with reduced feature count, combined feature groups and the created ensembles were evaluated using 10-fold cross validation, where after each iteration, the trained models are discarded. To reduce the effect of sample order, the cross validation process was repeated 100 times with randomised sample order. To keep training data as balanced as possible, each fold had equal number of healthy and depressed subjects. In case of predictions for the weighted and boosted ensemble, the training set in each fold undergoes additional 9-fold procedure (see Fig. \ref{fig:cv_chart}), to obtain prediction results for all samples in the training fold, afterwards the weights of classifier votes are fitted according to results in the training set. Similarly, AdaBoost used predicted class results from training set to calculate weights for each of the classifiers in ensemble. 
 
\subsection{Feature selection}
It is known that cognitive disorders can introduce observable change in measured EEG recordings. Depending on the used feature calculations, each brain region might have statistically significant difference when compared to cognitively normal patients brain. Therefore, to select the most relevant electrode locations we used feature subset selection methods that are applied in a prepossessing step before the model is learned. In particular, we used F-test which is widely used for showing statistical significance between two classes and ReliefF which is rank based feature selector.

\subsubsection{Univariate feature ranking using F-Tests}
The univariate feature ranking algorithm helps to understand the significance of each feature by examining the importance of each predictor individually using an F-test. Each F-test tests the hypothesis that the response values grouped by predictor variable values are drawn from populations with the same mean against the alternative hypothesis that the population means are different \cite{mathworkswebsite}.

\subsubsection{ReliefF}
The main idea of ReliefF algorithm is to estimate the quality of attributes on the basis of how well these attributes can distinguish between instances that are near to each other. The algorithm penalizes the predictors that give different values to neighbors of the same class, and rewards predictors that give different values to neighbors of different classes \cite{robnik2003theoretical}.

\subsection{Machine learning algorithms}
The supervised learning algorithms used in this study have been widely used in various EEG classification tasks according to survey papers published by Lakshmi et al \cite{lakshmi2014survey} and Alotaiby et al \cite{alotaiby2014eeg}. This study uses the fallowing algorithms for binary classification:

\begin{itemize}
  \item Support vector machine (SVM)
  \item Linear discriminant analysis (LDA) 
  \item Naive-byes (NB)
  \item K-nearest neighbours (kNN)
  \item Decision tree (D3)
\end{itemize}

In addition to individually evaluating the results for the listed classifiers and feature groups, an ensemble approach was also implemented, where classifiers trained on all 10 feature groups vote to predict the class label. 

\subsubsection{Support vector machine}
Support vector machine analyzes input data and recognizes patterns in a multi-dimensional feature space called the hyper-plane. An SVM model represents data as points in space, mapped so that the samples of the separate categories are clearly separable. New samples are then mapped into that same space and predicted to belong to a category based on the previously established hyper-planes \cite{cristianini2000introduction,subasi2010eeg, bhardwaj2015classification}.

This study shows the results for SVM with radial basis linear kernel function for two-class classification. The SVM used weighted standardizing where the software centers and scales each predictor variable by the corresponding weighted column mean and standard deviation.

\subsubsection{Linear discriminant analysis}
Discriminant analysis is a classification method that assumes that class features have different Gaussian distributions. To train a classifier, the fitting function estimates the parameters of a Gaussian distribution for each class in such a way as to minimize the variance and maximize the distance between the means of the two classes. To predict a class, the data is projected into LDA feature space and the trained model finds the class with the smallest misclassification cost \cite{fisher1936use,subasi2010eeg, bhardwaj2015classification}.

\subsubsection{Naive-byes classifier}
A Naive Bayes classifier which is based on Bayes’ Theorem assumes that the presence of a particular feature in a class is unrelated to the presence of any other feature \cite{hastie2009elements,sharmila2016dwt}. 

\subsubsection{K-nearest neighbor algorithm}
K-nearest neighbors algorithm uses feature similarity to predict the values of new samples based on how similar they matches the points in the training set. Similarity is usually measured as distance between points, where the Euclidean distance is one of the most popular choices \cite{duda2006pattern, sharmila2016dwt}. 

\subsubsection{Decision tree} 
Decision trees classifiers are structured similar to a tree structure where instances are classified according to their feature values. Decision trees classify instances by sorting them down the tree from the root to some leaf node, which provides the classification of the instance. An instance is classified by starting at the root node of the tree, testing the attribute specified by this node, then moving down the tree branch corresponding to the value of the attribute \cite{breiman1984ra, sun2007experimental}.

\subsubsection{Ensemble methods} 
An ensemble of classifiers is a method where several classifier decisions are combined in a manner that allows voting for classification new examples. Given that misclassified predictions are uncorrelated, the ensemble can correct for errors made by any individual classifier, leading to better overall accuracy \cite{ali1996error, abualsaud2015ensemble, datta2019comparative}.

The implemented ensemble votes are weighted according to majority voting where all weights are equal and weighted voting where weights are set according to classifier test set accuracy which was obtained by fallowing the procedure shown in Fig. \ref{fig:cv_chart}. The ensemble assigns $Label$ to a given sample according to fallowing equations 
\begin{equation}
 y = \sum_{n=1}^{m} w_i d_i
\end{equation}
where $m$ indicates number of classifiers, $w_i$ is the classifier weight and $d_i$ is the classifier decision $[1 = depressed$, $-1 = healthy]$. The class label is decided as fallows
\begin{equation}
Label=
\begin{cases}
  1, & \textit{if}\ y > 0 \\
  -1, & \textit{otherwise} 
\end{cases}
\end{equation}

As a third ensemble method we chose AdaBoost \cite{freund1997decision} (short for Adaptive Boosting) to see if it is possible to find more optimal weight combination in comparison to majority and weighted voting. The aim of AdaBoost is to convert a set of weak classifiers into a strong one. In our case, the set of weak classifiers are classifiers for each feature group. As with weighted voting, AdaBoost used predictions from the training set to find the most optimal classifier weights. 

\section{Results and discussion}
The baseline accuracy was established by individually evaluating all feature groups. In table \ref{tab:t1} the results for classifiers reached acceptable accuracy, where HFD and DFA reached above 80\% with at least one of classifiers. In all feature groups at least one classifier reached above 50\%, which indicates that feature groups contain relevant information and can be considered as week classifiers. Even tough the used features have been used successfully as depression biomarkers \cite{de2019depression}, some classifiers performed poorly, the assumption was that this happened because of redundant features, which representing electrode locations, harmed classifier performance.

\begin{table}[ht!]
\caption{Classifier accuracy }
\centering
\small
 \begin{tabular}{| c | c c c c c |} 
 \cline{2-6}
  \multicolumn{1}{c|}{} &  \multicolumn{5}{c|}{Classifier accuracy (\%)}\\
 \hline
  \shortstack{Feature \\ group} & \shortstack{rbf \\ SVM} & LDA & \shortstack{naive \\ Bayes} & kNN & D3\\
 \hline\hline
 T\textsubscript{RBP} & 54.40 & 65.00 & \textbf{73.30} & 44.45 & 38.65 \\ 
 \hline
 A\textsubscript{RBP} & 50.00 & 64.15 & \textbf{70.95} & 58.70 & 66.55 \\ 
 \hline
 B\textsubscript{RBP} & 70.00 & 52.90 & 65.05 & 62.20 & \textbf{79.85}\\ 
 \hline
 G\textsubscript{RBP} & 38.45 & 52.95 & \textbf{59.40} & 50.90 & 54.85\\ 
 \hline
 APV & 35.40 & 27.05 & 31.85 & 37.70 & \textbf{64.65}\\
 \hline
 SASI & 54.55 & 55.00 & 54.60 & \textbf{59.10} & 53.15\\ 
 \hline
 HFD & 55.40 & 41.55 & 51.80 & 71.55 & \textbf{82.55} \\ 
 \hline
 LZC & \textbf{80.70} & 57.10 & 58.50 & 75.95 & 63.50 \\  
 \hline
 DFA & 68.15 & 45.75 & 63.15 & 70.80 & \textbf{74.55} \\ 
 \hline
\end{tabular}
\label{tab:t1}
\end{table}

To address the problems with redundant features univariate feature ranking using F-Tests was used to select the most useful features. During classifier evaluation, classification results were obtained for all feature groups where features were sorted in order from most to least relevant. The evaluation started with the most relevant feature and in each iteration a less relevant feature was added to the feature set. The results presented in table \ref{tab:t2} show the highest accuracy achieved for particular number of features and the list of selected features can be found in table \ref{tab:t4}. The most optimal feature set was selected according to highest average accuracy of all five classifiers for each feature group. When compared to results from table \ref{tab:t1} to table \ref{tab:t2} the average classifier accuracy increased for almost all feature groups with the exception of LZC.

\begin{table}[ht!]
\caption{Classifier accuracy for features selected by univariate feature ranking using F-Tests}
\centering
\small
 \begin{tabular}{| c | c c c c c |} 
 \cline{2-6}
  \multicolumn{1}{c|}{} &  \multicolumn{5}{c|}{Classifier accuracy (\%)}\\
 \hline
  \shortstack{Feature \\ group} & \shortstack{rbf \\ SVM} & LDA & \shortstack{naive \\ Bayes} & kNN & D3\\
 \hline\hline
 T\textsubscript{RBP} (2) & 55.20 & 65.60 & \textbf{69.85} & 55.75 & 45.70 \\ 
 \hline
 A\textsubscript{RBP} (6) & 67.80 & 62.45 & \textbf{65.50} & 57.30 & 75.05 \\ 
 \hline
 B\textsubscript{RBP} (2) & \textbf{90.00} & 80.85 & 79.90 & \textbf{90.00} & \textbf{90.00}\\ 
 \hline
 G\textsubscript{RBP} (10) & 51.25 & 60.09 & \textbf{69.90} & 55.00 & 55.80\\ 
 \hline
 APV (7) & 34.90 & 36.30 & 33.30 & 46.25 & \textbf{73.05}\\
 \hline
 SASI (3) & \textbf{67.25} & 65.00 & 60.90 & 48.25 & 58.25\\ 
 \hline
 HFD (5) & \textbf{88.00} & 41.50 & 48.70 & 75.00 & 71.10 \\ 
 \hline
 LZC (10) & 60.65 & 67.50 & 56.25 & \textbf{75.00} & 68.15 \\  
 \hline
 DFA (3) & 66.90 & 64.45 & 57.90 & \textbf{68.65} & 67.85 \\ 
 \hline
 \multicolumn{4}{l}{(*) Number of features used}
\end{tabular}
\label{tab:t2}
\end{table}

Similarly as with F-tests, ReliefF was used to select the most relevant features for each feature group. Compared to the baseline and F-test feature selection classification results from tables \ref{tab:t1} and \ref{tab:t2}, the average classification results after ReliefF feature selection were higher for almost all feature groups (see table \ref{tab:t3}), results fro B\textsubscript{RBP} stayed the same because both feature selection algorithms had the best results using O1 and O2 electrodes.  

\begin{table}[ht!]
\caption{Classifier accuracy for features selected by ReliefF algorithm}
\centering
\small
 \begin{tabular}{| c | c c c c c |} 
 \cline{2-6}
  \multicolumn{1}{c|}{} &  \multicolumn{5}{c|}{Classifier accuracy (\%)}\\
 \hline
  \shortstack{Feature \\ group} & \shortstack{rbf \\ SVM} & LDA & \shortstack{naive \\ Bayes} & kNN & D3\\
 \hline\hline
 T\textsubscript{RBP} (5) & 66.15 & 79.60 & \textbf{80.00} & 72.25 & 55.85 \\ 
 \hline
 A\textsubscript{RBP} (2) & 81.20 & 78.70 & 75.95 & \textbf{90.00} & 85.30 \\ 
 \hline
 B\textsubscript{RBP} (2) & \textbf{90}.00 & 80.85 & 79.90 & \textbf{90.00} & \textbf{90.00}\\ 
 \hline
 G\textsubscript{RBP} (1) & \textbf{75.00} & 69.00 & \textbf{75.00} & 70.00 & 63.25\\ 
 \hline
 APV (3) & 58.85 & 51.55 & 48.65 & 60.60 & \textbf{65.85}\\
 \hline
 SASI (2) & 63.35 & 72.35 & 70.70 & 55.00 & \textbf{72.55}\\ 
 \hline
 HFD (4) & 78.75 & 49.35 & 65.65 & 75.95 & \textbf{85.70} \\ 
 \hline
 LZC (2) & 81.25 & 78.25 & 72.75 & \textbf{81.95} & 69.55 \\  
 \hline
 DFA (3) & 78.80 & 56.40 & 72.55 & \textbf{86.00} & 72.90 \\ 
 \hline
 \multicolumn{4}{l}{(*) Number of features used}
\end{tabular}
\label{tab:t3}
\end{table}

The selected features which were used for classification tasks mentioned in table \ref{tab:t2} for F-tests and table \ref{tab:t3} for ReliefF are shown in table \ref{tab:t4}. The selected features which represent electrode locations, show that the most used features are FP1, FP2 which correspond to frontal lobes and O1 and O2 at occipital lobes of the brain.

\begin{table}[ht!]
\caption{Selected channels based on F-test and ReliefF}
\centering
\small
 \begin{tabular}{| c | c | c |} 
  \cline{2-3}
  \multicolumn{1}{c|}{} &  \multicolumn{2}{c|}{Selected features}\\
 
 \hline
  \shortstack{Feature \\ group} & \shortstack{Univariate feature \\ ranking using F-Tests}  & ReliefF \\
 \hline\hline
 T\textsubscript{RBP} & F4 F8 & O1 PZ P4 C4 P3\\ 
 \hline
 A\textsubscript{RBP} & F4 F8 C3 T4 P4 O1 & O1 O2\\ 
 \hline
 B\textsubscript{RBP} & O1 O2 & O2 O1\\ 
 \hline
 G\textsubscript{RBP} & \shortstack{F7 F3 FZ F4 T3 \\ T4 O1 O2 FP1 FP2 } & FP1 \\ 
 \hline
 APV & T3 C3 F3 F4 F8 13 FP1 & F3 O2 C3\\
 \hline
 SASI & FP1 FP2 F7 & FP1 F3\\ 
 \hline
 HFD & FP1 FP2 FZ F8 C3 & FP1 O1 FP2 T5\\ 
 \hline
 LZC & \shortstack{F3 F4 T4 FP1 FP2 \\ FZ F8 P3 PZ O1} & FP1 FZ\\  
 \hline
 DFA & O1 O2 FP2 & FP1 FP2 O1\\ 
 \hline
\end{tabular}
\label{tab:t4}
\end{table}

Instead of focusing on classification of individual feature groups, combined features where also evaluated, where the feature vector is created by concatenating all available features. In case of selected features, only the features listed in table \ref{tab:t4} are concatenated. Table \ref{tab:t5} clearly shows the benefit of feature selection. For the most part the classification results for feature selected based on F-test and ReliefF are higher than the baseline results for selecting all features.

\begin{table}[ht!]
\caption{Concatenated feature classifier accuracy }
\centering
\small
 \begin{tabular}{| c | c c c c c |} 
 \cline{2-6}
  \multicolumn{1}{c|}{} &  \multicolumn{5}{c|}{Classifier accuracy (\%)}\\
 \hline
  \shortstack{Features} & \shortstack{rbf \\ SVM} & LDA & \shortstack{naive \\ Bayes} & kNN & D3\\
 \hline\hline
 All features & 52.40 & 51.35 & \textbf{59.70} & 55.50 & 53.45 \\ 
  F-test features & 52.10 & 59.65 & 66.25 & 62.15 & \textbf{73.75} \\ 
   ReliefF features & 55.10 & 71.75 & 72.35 & \textbf{80.00} & 62.35 \\ 
 \hline
 
\end{tabular}
\label{tab:t5}
\end{table}

A more robust solution can be achieved using an ensemble approach where many weak classifiers contribute to the classification prediction by voting. Each result showed in table \ref{tab:t6} is the result of combining 10 classifiers of the same type; one classifier for each feature group. The features used in each feature group were selected according to table \ref{tab:t4}. The majority voting ensemble approach shows significant improvement compared to using all features with one classifier. When compared to results in table \ref{tab:t5}, ensemble approach further improves results when using F-tests and ReliefF feature selection algorithms \ref{tab:t5}. On average ReliefF classification results outperformed ensembles whose features were selected by F-tests.

The use of AdaBoost for classifier weight selection in most of the cased did not improve results compared to majority voting and weighted ensemble. Due to the nature of AdaBoost, during the weight calculation process, the algorithm can reach optimal weights using only few of the classifiers and ignore the rest, which hinders the robustness of the ensemble.   

\begin{table}[ht!]
\caption{Ensemble classifier accuracy}
\centering
\small
 \begin{tabular}{| c | c c c c c |} 
 \cline{2-6}
  \multicolumn{1}{c|}{} &  \multicolumn{5}{c|}{Classifier accuracy (\%)}\\
 \hline
  \shortstack{Feature group \\ ensemble} & \shortstack{rbf \\ SVM} & LDA & \shortstack{naive \\ Bayes} & kNN & D3\\
 \hline\hline
  All + Majority & 70.45 & 58.00 & 61.15 & 68.05 & \textbf{76.15} \\ 
  F-test + Majority & 80.85 & 70.50 & \textbf{81.15} & 77.05 & 78.70  \\ 
 ReliefF + Majority & 85.40 & 77.55 & \textbf{90.95} & 83.80 & 90.10\\ 
 \hline
  All + Weighted & 65.45 & 56.45 & 63.75 & 68.85 & \textbf{72.60} \\ 
  F-test + Weighted & \textbf{81.50} & 69.70 & 78.10 & 79.50 & 78.40 \\ 
 ReliefF + Weighted & 84.10 & 79.20 & 85.85 & 87.95 & \textbf{88.30}\\ 
 \hline
 All + Adaboost & 70.80 & 56.55 & \textbf{71.50} & 63.75 & 68.60 \\ 
 F-test + Adaboost & 80.60 & 70.00 & 78.90 & \textbf{83.60} & 79.55\\ 
 ReliefF + Adaboost & 79.90 & 72.45 & 78.45 & \textbf{84.45} & 81.00\\ 
 \hline
\end{tabular}
\label{tab:t6}
\end{table}

\section{Conclusion}

This study results indicate the potential use of EEG relative band power, linear (APV, RGP, SASI) and non-linear (HDF, LZC, DFA) EEG features, which have been successfully used for classification and regression tasks in detecting depression, for classification of long lasting effects of depression. 

The described feature groups and classification methods, which were evaluated using 10-fold cross validation were used to classify age and gender matched 10 healthy subjects and 10 subjects who have had depression with up to 82\% accuracy with features groups such as HDF and LZC using D3 and rbf SVM binary classifier. The results improve using feature selection algorithms such as univariate feature ranking using F-Tests and ReliefF feature selection algorithm which improved the classification accuracy up to 90\%. The mentioned results refer to results of individual feature groups in combination with a single classifier. In addition, in a majority voting and weighted ensemble setup, the accuracy of used classifiers (SVM, LDA, NB, kNN, D3) show consistently high results with highest accuracy of 90.95\%. 

\section*{Acknowledgements}This work is supported by the Estonian Centre of Excellence in IT (EXCITE) funded by the European Regional Development Fund. The authors also gratefully acknowledge the support of NVIDIA Corporation with the donation of the Titan XP Pascal GPU.

\printbibliography

@article{aluoja1999development,
  title={Development and psychometric properties of the Emotional State Questionnaire, a self-report questionnaire for depression and anxiety},
  author={Aluoja, Anu and Shlik, Jakov and Vasar, Veiko and Luuk, Kersti and Leinsalu, Mall},
  journal={Nordic Journal of Psychiatry},
  volume={53},
  number={6},
  pages={443--449},
  year={1999},
  publisher={Taylor \& Francis}
}

@article{mahato2019detection,
  title={Detection of major depressive disorder using linear and non-linear features from EEG signals},
  author={Mahato, Shalini and Paul, Sanchita},
  journal={Microsystem Technologies},
  volume={25},
  number={3},
  pages={1065--1076},
  year={2019},
  publisher={Springer}
}

@article{park2007multiscale,
  title={Multiscale entropy analysis of EEG from patients under different pathological conditions},
  author={Park, Jeong-Hyeon and Kim, Sooyong and Kim, Cheol-Hyun and Cichocki, Andrzej and Kim, Kyungsik},
  journal={Fractals},
  volume={15},
  number={04},
  pages={399--404},
  year={2007},
  publisher={World Scientific}
}

@article{ibanez2015multiscale,
  title={Multiscale Lempel--Ziv complexity for EEG measures},
  author={Ibanez-Molina, Antonio J and Iglesias-Parro, Sergio and Soriano, Maria F and Aznarte, Jose I},
  journal={Clinical Neurophysiology},
  volume={126},
  number={3},
  pages={541--548},
  year={2015},
  publisher={Elsevier}
}

@article{hinrikus2009electroencephalographic,
  title={Electroencephalographic spectral asymmetry index for detection of depression},
  author={Hinrikus, Hiie and Suhhova, Anna and Bachmann, Maie and Aadamsoo, Kaire and V{\~o}hma, {\"U}lle and Lass, Jaanus and Tuulik, Viiu},
  journal={Medical \& biological engineering \& computing},
  volume={47},
  number={12},
  pages={1291},
  year={2009},
  publisher={Springer}
}

@inproceedings{kalev2015lempel,
  title={Lempel-Ziv and multiscale Lempel-Ziv complexity in depression},
  author={Kalev, Kaia and Bachmann, Maie and Orgo, L and Lass, Jaanus and Hinrikus, Hiie},
  booktitle={2015 37th Annual International Conference of the IEEE Engineering in Medicine and Biology Society (EMBC)},
  pages={4158--4161},
  year={2015},
  organization={IEEE}
}

@inproceedings{bachmann2014detrended,
  title={Detrended fluctuation analysis of EEG in depression},
  author={Bachmann, M and Suhhova, A and Lass, J and Aadamsoo, K and V{\~o}hma, {\"U} and Hinrikus, H},
  booktitle={XIII Mediterranean Conference on Medical and Biological Engineering and Computing 2013},
  pages={694--697},
  year={2014},
  organization={Springer}
}

@article{bachmann2017single,
  title={Single channel EEG analysis for detection of depression},
  author={Bachmann, Maie and Lass, Jaanus and Hinrikus, Hiie},
  journal={Biomedical Signal Processing and Control},
  volume={31},
  pages={391--397},
  year={2017},
  publisher={Elsevier}
}

@article{bachmann2018methods,
  title={Methods for classifying depression in single channel EEG using linear and nonlinear signal analysis},
  author={Bachmann, Maie and P{\"a}eske, Laura and Kalev, Kaia and Aarma, Katrin and Lehtmets, Andres and {\"O}{\"o}pik, Pille and Lass, Jaanus and Hinrikus, Hiie},
  journal={Computer methods and programs in biomedicine},
  volume={155},
  pages={11--17},
  year={2018},
  publisher={Elsevier}
}

@article{bachmann2013spectral,
  title={Spectral asymmetry and Higuchi’s fractal dimension measures of depression electroencephalogram},
  author={Bachmann, Maie and Lass, Jaanus and Suhhova, Anna and Hinrikus, Hiie},
  journal={Computational and mathematical methods in medicine},
  volume={2013},
  year={2013},
  publisher={Hindawi}
}

@article{wittchen2011size,
  title={The size and burden of mental disorders and other disorders of the brain in Europe 2010},
  author={Wittchen, Hans-Ulrich and Jacobi, Frank and Rehm, J{\"u}rgen and Gustavsson, Anders and Svensson, Mikael and J{\"o}nsson, Bengt and Olesen, Jes and Allgulander, Christer and Alonso, Jordi and Faravelli, Carlo and others},
  journal={European neuropsychopharmacology},
  volume={21},
  number={9},
  pages={655--679},
  year={2011},
  publisher={Elsevier}
}

@article{cai2018pervasive,
  title={A pervasive approach to EEG-based depression detection},
  author={Cai, Hanshu and Han, Jiashuo and Chen, Yunfei and Sha, Xiaocong and Wang, Ziyang and Hu, Bin and Yang, Jing and Feng, Lei and Ding, Zhijie and Chen, Yiqiang and others},
  journal={Complexity},
  volume={2018},
  year={2018},
  publisher={Hindawi}
}

@article{beck1996manual,
  title={Manual for the Beck depression inventory-II. 1996},
  author={Beck, AT and Steer, RA and Brown, GK},
  journal={San Antonio, TX: Psychological Corporation},
  volume={2},
  year={1996}
}

@article{mowbray1972hamilton,
  title={The Hamilton Rating Scale for depression: a factor analysis},
  author={Mowbray, RM},
  journal={Psychological Medicine},
  volume={2},
  number={3},
  pages={272--280},
  year={1972},
  publisher={Cambridge University Press}
}

@article{tombaugh1996mini,
  title={Mini-Mental State Examination (MMSE) and the Modified MMSE (3MS): a psychometric comparison and normative data.},
  author={Tombaugh, TN and McDowell, I and Kristjansson, B and Hubley, AM},
  journal={Psychological Assessment},
  volume={8},
  number={1},
  pages={48},
  year={1996},
  publisher={American Psychological Association}
}

@article{lehman2000diagnostic,
  title={The diagnostic and statistical manual of mental disorders},
  author={Lehman, Jill Fain},
  year={2000},
  publisher={Citeseer}
}

@book{murray1996global,
  title={The global burden of disease: a comprehensive assessment of mortality and disability from diseases, injuries, and risk factors in 1990 and projected to 2020: summary},
  author={Murray, Christopher JL and Lopez, Alan D and World Health Organization and others},
  year={1996},
  publisher={World Health Organization}
}

@article{lakshmi2014survey,
  title={Survey on EEG signal processing methods},
  author={Lakshmi, M Rajya and Prasad, TV and Prakash, Dr V Chandra},
  journal={International Journal of Advanced Research in Computer Science and Software Engineering},
  volume={4},
  number={1},
  year={2014}
}

@article{knott2001eeg,
  title={EEG power, frequency, asymmetry and coherence in male depression},
  author={Knott, Verner and Mahoney, Colleen and Kennedy, Sidney and Evans, Kenneth},
  journal={Psychiatry Research: Neuroimaging},
  volume={106},
  number={2},
  pages={123--140},
  year={2001},
  publisher={Elsevier}
}

@article{khosla2020comparative,
  title={A comparative analysis of signal processing and classification methods for different applications based on EEG signals},
  author={Khosla, Ashima and Khandnor, Padmavati and Chand, Trilok},
  journal={Biocybernetics and Biomedical Engineering},
  year={2020},
  publisher={Elsevier}
}

@article{de2019depression,
  title={Depression biomarkers using non-invasive EEG: A review},
  author={de Aguiar Neto, Fernando Soares and Rosa, Jo{\~a}o Luis Garcia},
  journal={Neuroscience \& Biobehavioral Reviews},
  volume={105},
  pages={83--93},
  year={2019},
  publisher={Elsevier}
}

@article{zhang2001eeg,
  title={EEG complexity as a measure of depth of anesthesia for patients},
  author={Zhang, X-S and Roy, Rob J and Jensen, Erik W},
  journal={IEEE transactions on biomedical engineering},
  volume={48},
  number={12},
  pages={1424--1433},
  year={2001},
  publisher={IEEE}
}

@incollection{mahato2019electroencephalogram,
  title={Electroencephalogram (EEG) signal analysis for diagnosis of major depressive disorder (MDD): a review},
  author={Mahato, Shalini and Paul, Sanchita},
  booktitle={Nanoelectronics, Circuits and Communication Systems},
  pages={323--335},
  year={2019},
  publisher={Springer}
}

@article{peng1994mosaic,
  title={Mosaic organization of DNA nucleotides},
  author={Peng, C-K and Buldyrev, Sergey V and Havlin, Shlomo and Simons, Michael and Stanley, H Eugene and Goldberger, Ary L},
  journal={Physical review e},
  volume={49},
  number={2},
  pages={1685},
  year={1994},
  publisher={APS}
}

@article{peng1995quantification,
  title={Quantification of scaling exponents and crossover phenomena in nonstationary heartbeat time series},
  author={Peng, C-K and Havlin, Shlomo and Stanley, H Eugene and Goldberger, Ary L},
  journal={Chaos: an interdisciplinary journal of nonlinear science},
  volume={5},
  number={1},
  pages={82--87},
  year={1995},
  publisher={American Institute of Physics}
}

@article{higuchi1988approach,
  title={Approach to an irregular time series on the basis of the fractal theory},
  author={Higuchi, Tomoyuki},
  journal={Physica D: Nonlinear Phenomena},
  volume={31},
  number={2},
  pages={277--283},
  year={1988},
  publisher={Elsevier}
}

@article{robnik2003theoretical,
  title={Theoretical and empirical analysis of ReliefF and RReliefF},
  author={Robnik-{\v{S}}ikonja, Marko and Kononenko, Igor},
  journal={Machine learning},
  volume={53},
  number={1-2},
  pages={23--69},
  year={2003},
  publisher={Springer}
}

@article{alotaiby2014eeg,
  title={EEG seizure detection and prediction algorithms: a survey},
  author={Alotaiby, Turkey N and Alshebeili, Saleh A and Alshawi, Tariq and Ahmad, Ishtiaq and Abd El-Samie, Fathi E},
  journal={EURASIP Journal on Advances in Signal Processing},
  volume={2014},
  number={1},
  pages={183},
  year={2014},
  publisher={Springer}
}

@online{mathworkswebsite,
    title = "Univariate feature ranking for regression using F-tests",
    url  = "https://se.mathworks.com/help/stats/fsrftest.html",
    addendum = "(accessed: 01.01.2021)"
}

@article{subasi2010eeg,
  title={EEG signal classification using PCA, ICA, LDA and support vector machines},
  author={Subasi, Abdulhamit and Gursoy, M Ismail},
  journal={Expert systems with applications},
  volume={37},
  number={12},
  pages={8659--8666},
  year={2010},
  publisher={Elsevier}
}

@inproceedings{bhardwaj2015classification,
  title={Classification of human emotions from EEG signals using SVM and LDA Classifiers},
  author={Bhardwaj, Aayush and Gupta, Ankit and Jain, Pallav and Rani, Asha and Yadav, Jyoti},
  booktitle={2015 2nd International Conference on Signal Processing and Integrated Networks (SPIN)},
  pages={180--185},
  year={2015},
  organization={IEEE}
}

@article{sharmila2016dwt,
  title={DWT based detection of epileptic seizure from EEG signals using naive Bayes and k-NN classifiers},
  author={Sharmila, A and Geethanjali, P},
  journal={Ieee Access},
  volume={4},
  pages={7716--7727},
  year={2016},
  publisher={IEEE}
}

@article{sun2007experimental,
  title={An experimental evaluation of ensemble methods for EEG signal classification},
  author={Sun, Shiliang and Zhang, Changshui and Zhang, Dan},
  journal={Pattern Recognition Letters},
  volume={28},
  number={15},
  pages={2157--2163},
  year={2007},
  publisher={Elsevier}
}

@incollection{datta2019comparative,
  title={Comparative study of different ensemble compositions in eeg signal classification problem},
  author={Datta, Ankita and Chatterjee, Rajdeep},
  booktitle={Emerging Technologies in Data Mining and Information Security},
  pages={145--154},
  year={2019},
  publisher={Springer}
}

@article{freund1997decision,
  title={A decision-theoretic generalization of on-line learning and an application to boosting},
  author={Freund, Yoav and Schapire, Robert E},
  journal={Journal of computer and system sciences},
  volume={55},
  number={1},
  pages={119--139},
  year={1997},
  publisher={Elsevier}
}

@book{duda2006pattern,
  title={Pattern classification},
  author={Duda, Richard O and Hart, Peter E and others},
  year={2006},
  publisher={John Wiley \& Sons}
}

@article{breiman1984ra,
  title={RA Olshen and CJ Stone,“},
  author={Breiman, L and Friedman, JH},
  journal={Classification and regression trees},
  year={1984}
}

@book{hastie2009elements,
  title={The elements of statistical learning: data mining, inference, and prediction},
  author={Hastie, Trevor and Tibshirani, Robert and Friedman, Jerome},
  year={2009},
  publisher={Springer Science \& Business Media}
}

@article{fisher1936use,
  title={The use of multiple measurements in taxonomic problems},
  author={Fisher, Ronald A},
  journal={Annals of eugenics},
  volume={7},
  number={2},
  pages={179--188},
  year={1936},
  publisher={Wiley Online Library}
}

@book{cristianini2000introduction,
  title={An introduction to support vector machines and other kernel-based learning methods},
  author={Cristianini, Nello and Shawe-Taylor, John and others},
  year={2000},
  publisher={Cambridge university press}
}

@article{abualsaud2015ensemble,
  title={Ensemble classifier for epileptic seizure detection for imperfect EEG data},
  author={Abualsaud, Khalid and Mahmuddin, Massudi and Saleh, Mohammad and Mohamed, Amr},
  journal={The Scientific World Journal},
  volume={2015},
  year={2015},
  publisher={Hindawi}
}

@article{ali1996error,
  title={Error reduction through learning multiple descriptions},
  author={Ali, Kamal M and Pazzani, Michael J},
  journal={Machine learning},
  volume={24},
  number={3},
  pages={173--202},
  year={1996},
  publisher={Springer}
}

@article{lempel1976complexity,
  title={On the complexity of finite sequences},
  author={Lempel, Abraham and Ziv, Jacob},
  journal={IEEE Transactions on information theory},
  volume={22},
  number={1},
  pages={75--81},
  year={1976},
  publisher={IEEE}
}

\end{document}